\documentclass[sigconf,authorversion,breaklinks=true]{acmart}
\usepackage[utf8]{inputenc}
\usepackage[T1]{fontenc}
\usepackage{amsmath}
\usepackage{graphicx}
\usepackage{fancyhdr}
\usepackage{breakcites}
\usepackage[export]{adjustbox}

\usepackage{multirow}
\usepackage{subfig}

\usepackage{longtable}
\usepackage{booktabs}
\usepackage{todonotes}
\usepackage{enumitem}
\usepackage{lscape}
\usepackage{float}
\usepackage{geometry}

\AtBeginDocument{%
  \providecommand\BibTeX{{%
    \normalfont B\kern-0.5em{\scshape i\kern-0.25em b}\kern-0.8em\TeX}}}

\copyrightyear{2020}
\acmYear{2020}
\setcopyright{acmlicensed}
\acmConference[ACM CHIL '20]{ACM Conference on Health, Inference, and Learning}{April 2--4, 2020}{Toronto, ON, Canada}
\acmBooktitle{ACM Conference on Health, Inference, and Learning (ACM CHIL '20), April 2--4, 2020, Toronto, ON, Canada}
\acmPrice{15.00}
\acmDOI{10.1145/3368555.3384469}
\acmISBN{978-1-4503-7046-2/20/04}

\begin{document}

\title{MIMIC-Extract: A Data Extraction, Preprocessing, and Representation Pipeline for MIMIC-III}



\author{Shirly Wang}
\authornote{Both authors contributed equally to this research.}
\affiliation{%
    \institution{Layer 6 AI, University of Toronto}
}
\email{shirlywang@cs.toronto.edu}

\author{Matthew B.A. McDermott}
\authornotemark[1]
\affiliation{%
    \institution{Massachusetts Institute of Technology}
}
\email{mmd@mit.edu}

\author{Geeticka Chauhan}
\affiliation{%
    \institution{Massachusetts Institute of Technology}
}
\email{geeticka@mit.edu}

\author{Marzyeh Ghassemi}
\affiliation{%
    \institution{University of Toronto, Vector Insitute}
}
\email{marzyeh@cs.toronto.edu}

\author{Michael C. Hughes}
\affiliation{%
    \institution{Tufts University}
}
\email{mhughes@cs.tufts.edu}

\author{Tristan Naumann}
\affiliation{%
    \institution{Microsoft Research}
}
\email{tristan@microsoft.com}

\renewcommand{\shortauthors}{S. Wang, M. B. A. McDermott, G. Chauhan, M. Ghassemi, M. C. Hughes, and T. Naumann}

\begin{abstract}
  Machine learning for healthcare researchers face challenges to progress and reproducibility due to a lack of standardized processing frameworks for public datasets. We present \texttt{MIMIC-Extract}, an open source pipeline for transforming the raw electronic health record (EHR) data of critical care patients from the publicly-available MIMIC-III database into data structures that are directly usable in common time-series prediction pipelines. \texttt{MIMIC-Extract} addresses three challenges in making complex EHR data accessible to the broader machine learning community. First, \texttt{MIMIC-Extract} transforms raw vital sign and laboratory measurements into usable hourly time series, performing essential steps such as unit conversion, outlier handling, and aggregation of semantically similar features to reduce missingness and improve robustness. Second, \texttt{MIMIC-Extract} extracts and makes prediction of clinically-relevant targets possible, including outcomes such as mortality and length-of-stay, as well as comprehensive hourly intervention signals for ventilators, vasopressors, and fluid therapies. Finally, the pipeline emphasizes reproducibility and is extensible to enable future research questions. We demonstrate the pipeline's effectiveness by developing several benchmark tasks for outcome and intervention forecasting and assessing the performance of competitive models.
\end{abstract}



\maketitle

\section*{Introduction}
Applying modern machine learning to observational health data holds the potential to improve healthcare in many ways, such as delivering better patient treatments, improving hospital operations, and answering fundamental scientific questions~\citep{ghassemi2018opportunities}. To realize this potential, there have been efforts to make healthcare data available to credentialed researchers with human subjects training. A widely-used public data source is the Medical Information Mart for Intensive Care (MIMIC-III) dataset~\citep{johnson2016mimic}, which makes available the de-identified electronic health records (EHRs) of 53,423 patients admitted to critical care units at a Boston-area hospital from 2001--2012. While MIMIC-III's availability has catalyzed many research studies, working with MIMIC-III data remains technically challenging, which presents a barrier to entry. The primary difficulties rest in the complexity of EHR data and the myriad choices that must be made to extract a clinically-relevant cohort for analysis. These same difficulties hinder the reproducibility of studies that apply machine learning to MIMIC-III data, because researchers develop code independently to extract and preprocess task-appropriate cohorts. The majority of papers do not share code used to extract study-specific data~\citep{johnson2017reproducibility}, resulting in expensive yet redundant efforts to build upon existing work and creating the potential for hard-to-explain differences in results.

\begin{figure}[t!]
    \includegraphics[width=\linewidth]{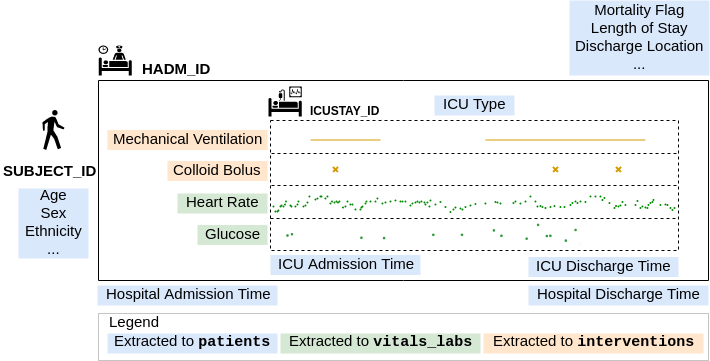}
    \caption{Example data produced by \texttt{MIMIC-Extract} to summarize a single subject's stay in the intensive care unit(ICU). Time evolves on the x-axis, and all extracted time series are discretized into hourly buckets.
    Mechanical Ventilation is an example intervention with multi-hour continuous duration. Colloid bolus is an example of an intermittent fluids intervention.
    All interventions are recorded as binary indicators at each hour.
    Heart Rate is an example of a frequent vital sign. Glucose is an example of an infrequent lab measurement.
    }
    \label{fig:data_extract}
\end{figure}

In this paper, we introduce \texttt{MIMIC-Extract},\footnote{\url{https://github.com/MLforHealth/MIMIC_Extract}} an open source pipeline to extract, preprocess, and represent data from MIMIC-III v1.4, including static demographic information available at admission, time-varying vital signs and laboratory measurements, time-varying intervention signals, and static outcomes such as length-of-stay or mortality.
Figure~\ref{fig:data_extract} gives a visual summary of the data we extract from the observed records of an individual patient stay available in MIMIC-III.
Our principled approach yields a comprehensive cohort of time-series data that is well-suited for several clinically-meaningful prediction tasks --- several of which we profile in this paper --- while simultaneously providing flexibility in cohort selection and variable selection.


We intend this pipeline to serve as a foundation for both benchmarking the state-of-the-art and enabling progress on new research tasks. Several other recent works have developed, in parallel, extraction pipelines and prediction benchmark tasks for MIMIC-III data~\citep{harutyunyan2017multitask,purushotham2018benchmark,sjoding2019democratizing}. However, compared to these we advance the field with three primary contributions:

\begin{itemize}[leftmargin=*]

\item \textbf{Robust Representations of Labs and Vitals Time Series.}
The primary difficulty of using the raw MIMIC-III data is the noisy nature of clincal data. 
We present a comprehensive procedure designed with clinical validity in mind to standardize units of measurement, detect and correct outliers, and select a curated set of features that reduce data missingness.
Importantly, we offer data representations that are \emph{resilient} to concept drift over time, by aggregating semantically similar raw features.
The robustness of this ``clinically aggregated'' representation is demonstrated by recent work on feature robustness in non-stationary health records~\citep{nestor2019}.


i
\item \textbf{Clinically Meaningful Interventions and Outcomes.}
Our pipeline focuses on making hourly-observed treatment signals available for several actionable critical care interventions, including ventilation, vasopressors (for blood pressure management), and fluid bolus therapies (for managing sepsis and other conditions). 
No other recent pipeline makes interventions a primary focus.
We also support several common outcomes of interest, such as mortality and length of stay.
We intentionally avoid tasks of questionable clinical utility appearing in some prior works, such as diagnosis billing code prediction, because they have poor diagnostic value~\citep{agniel2018biases}.
In later benchmark task design, we further emphasize realistic settings such as predictions that occur every hour rather than after a single 24-hour duration.
We are careful to include meaningful temporal gaps between measurement and outcome, in order to minimizes label leakage and thus improve the utility of models in real clinical deployment.
	
\item \textbf{Focus on Usability, Reproducibility, and Extensibility.}
Finally, we have designed the entire pipeline with usability and extensibility in mind.
Our patient selection criteria can be easily adjusted to support future research questions, requiring changes to only keyword arguments rather than source code.
Extracted data can be read directly into a Pandas DataFrame~\citep{mckinney2010data} with appropriate data typing, enabling immediate computational analysis.
We also provide Jupyter Notebooks~\citep{perez2007ipython} that demonstrate the use of the data produced by our \texttt{MIMIC-Extract} pipeline in benchmark prediction tasks, including steps for data loading and preprocessing, and baseline model building.
\end{itemize}

We emphasize that our pipeline has been used as the foundation for reproducing many recent machine learning studies of MIMIC-III data~\citep{ghassemi2014unfolding,ghassemi2015multivariate,wu2016ssam,ghassemi2017predicting,suresh2017clinical,raghu2017continuous,mcdermott2018semi,nestor2019}.
While none of these released their own extraction code, they nevertheless utilized similar cohort selection and variable selection processes.

The rest of this paper provides an overview of the extraction system design, a detailed comparison to other extraction systems and their corresponding benchmark tasks, and a careful analysis of several benchmark prediction tasks developed using our pipeline to showcase its potential.


\section*{Data Pipeline Overview}

\begin{figure}[h!]
    \includegraphics[width=\linewidth]{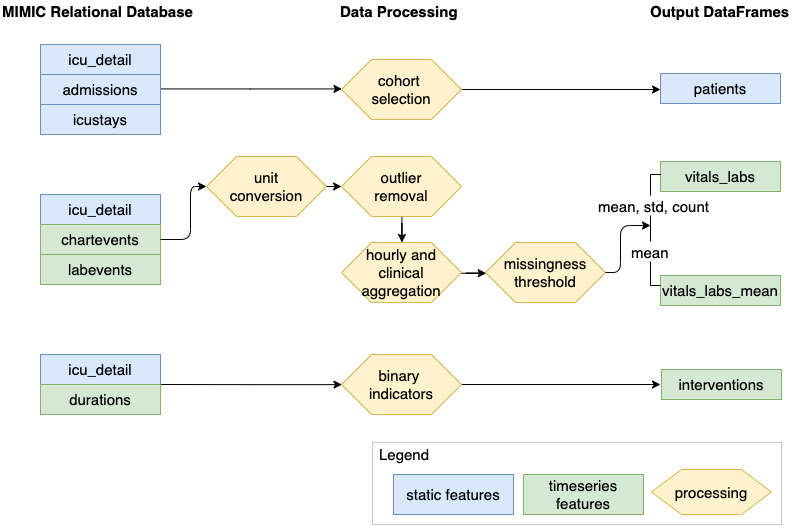}
    \caption{\texttt{MIMIC-Extract} Overview: First, a cohort is created that meets our selection criteria. Static demographic variables and ICU stay information for patients in the cohort are extracted and stored in \texttt{patients}. Next, labs and vitals for patients in the cohort are extracted and stored in \texttt{vital\_labs} and \texttt{vitals\_labs\_mean}. By default, only labs and vitals that are missing less frequently than a pre-defined threshold are extracted and outlier values are filtered based on physiological valid ranges. Finally, hourly intervention time series for the same patients are extracted and stored in \texttt{interventions}.}
    \label{fig:pipeline}
\end{figure}

Figure~\ref{fig:pipeline} summarizes the data extraction and processing steps involved in \texttt{MIMIC-Extract}. From the MIMIC relational database, SQL query results are processed to generate four output tables. These tables, as summarized in Table~\ref{tab:dataframe}, maintain the time series nature of clinical data and also provide an aggregated featurization of the cohort selected.

\begin{table*}[h]
\begin{small}
\centering
\resizebox{\linewidth}{!}{%
\begin{tabular}{lll} \toprule
Table Name                  & Index   & Variables  \\ \midrule
\texttt{patients}   & \texttt{subject\_id}, \texttt{hadm\_id}, \texttt{icustay\_id}                    & static demographics, static outcomes                                      \\
\texttt{vitals\_labs}       & \texttt{subject\_id}, \texttt{hadm\_id}, \texttt{icustay\_id}, \texttt{hours\_in} & time-varying vitals and labs (hourly mean, count and standard deviation)  \\
\texttt{vitals\_labs\_mean} & \texttt{subject\_id}, \texttt{hadm\_id}, \texttt{icustay\_id}, \texttt{hours\_in} & time-varying vitals and labs (hourly mean only)                           \\
\texttt{interventions}      & \texttt{subject\_id}, \texttt{hadm\_id}, \texttt{icustay\_id}, \texttt{hours\_in} & hourly binary indicators for administered interventions                   \\ 
\bottomrule
\end{tabular}}
\end{small}
\caption{Description of all output tables generated by \texttt{MIMIC-Extract}.}
\label{tab:dataframe}
\end{table*}

\subsection*{Cohort Selection}

The MIMIC-III database captures over a decade of intensive care unit (ICU) patient stays at Beth Israel Deaconess Medical Center.
An individual patient might be admitted to the ICU at multiple times over the years, and even within a single hospital stay could be moved in and out of the ICU multiple times. 
We choose to focus on each subject's first ICU visit only, since those who make repeat visits typically require additional considerations with respect to modeling and providing useful treatment.
Our proposed pipeline thus includes all patient ICU stays in the MIMIC-III database that meet the following criteria: the subject is an adult (age of at least 15 at time of admission), the stay is the first known ICU admission for the subject, and the total duration of the stay is at least 12 hours and less than 10 days.
This cohort selection is consistent with many previous papers using MIMIC-III~\cite{ghassemi2014unfolding,ghassemi2015multivariate,wu2016ssam,ghassemi2017predicting,suresh2017clinical,raghu2017continuous,mcdermott2018semi}.

\subsection*{Variable Selection}

\paragraph{Static Variables}
By default, our extraction code extracts all 10 static demographic variables listed in Table~\ref{tab:static_set}, along with static outcomes including in-ICU mortality, in-hospital mortality, and the patient's total ICU length-of-stay (LOS), in hours.
Our pipeline presents values for static variables as they originally appear in MIMIC-III raw data with no additional outlier removal. For example, age for patients older than eighty-nine is masked as 300 in MIMIC-III for privacy reasons, and our pipeline preserves this sentinel value to allow downstream handling of these subjects.

\begin{table}[h!]
\centering
\resizebox{\linewidth}{!}{%
\begin{tabular}{l|l} 
\toprule
\textbf{Variable}        & \textbf{Concept}                                                            \\ 
\midrule
\multirow{2}{*}{age}             & patient age (masked as 300 for patients \\
& older than 89 years old in MIMIC-III)  \\
ethnicity       & patient ethnicity                                                  \\
gender          & patient gender                                                     \\
insurance       & patient insurance type                                             \\
admittime       & hospital admission time                                            \\
dischtime       & hospital discharge time                                                     \\
intime          & ICU admission time                                                 \\
outtime         & ICU discharge time                                                 \\
admission\_type & type of hospital admission                                         \\
first\_careunit & type of ICU when first admitted                           \\
\bottomrule
\end{tabular}}
\caption{Static demographic variables and admission information generated by \texttt{MIMIC-Extract}.}
\label{tab:static_set}
\end{table}

\paragraph{Time-Varying Vitals and Labs}
By default, our extraction code extracts 104 clinically aggregated time-series variables (listed in Appendix~\ref{app:feature_set}) related to vital signs (e.g., heart rate or blood pressure) and laboratory test results (e.g., white blood cell counts).
These were selected as a comprehensive set of possible signals for prediction algorithms with input from clinical care teams. 
Practitioners can optionally choose to output only a subset of these variables that meet certain minimum percentages of non-missingness, as explained in later sections.

When comparing our selected features to previous work, we find that we include all 12 time-varying features in the small curated set of 17 features considered by~\citet{purushotham2018benchmark} (the other 5 include two static features we use, age and admission type, and three diagnosis code features we intentionally omit).
We include 13 of the 17 time-varying vitals and labs featured in \citet{harutyunyan2017multitask}'s recent pipeline (we omit capillary refill rate due to high missingness rates as do all the feature sets surveyed by~\citet{purushotham2018benchmark}; we further do not consider the separate eye, motor and verbal Glascow coma scores, only the total score). 
Importantly, unlike the large set of 136 ``raw'' features advocated by \citet{purushotham2018benchmark}, we do not include any prescription drugs such as aspirin --- this is an intentional omission, because of the unclear quality of the prescription signals in the MIMIC-III database.
Without additional insight into the prescriptions a patient actually took, which may differ from all prescriptions ordered for a patient, we feel the inclusion of prescriptions can induce significant confounding effects on the resulting models.

\subsection*{Unit Conversion and Outlier Detection}
Sometimes vitals and labs are recorded with different measuring units in EHR data. Our data pipeline standardizes measurements into consistent units, including weight into kilograms, height into centimeters, and temperature into degrees Celsius. This process is easily extensible if any additional unit-classes are added by downstream users which need conversion.

To handle outliers, we make use of a list of clinically reasonable variable ranges provided in the source code repository of \citet{harutyunyan2017multitask},\footnote{\url{https://github.com/YerevaNN/mimic3-benchmarks/blob/master/mimic3benchmark/resources/variable_ranges.csv}. Accessed 2019-03-29.} which was developed in conversation with clinical experts, based on their knowledge of valid clinical measure ranges.
Each numerical variable is associated with upper and lower thresholds for detecting unusable outliers. If the raw observed value falls outside these thresholds, it is treated as missing. Additionally, each variable is associated with more refined upper and lower thresholds for defining the \emph{physiologically valid} range of measurements. Any non-outlier value that falls outside the physiologically valid range is replaced with the nearest valid value. In generating the default cohort, we replace 35,251 (0.05\%) measurements classified as non-valid outliers with nearest valid values and remove 5,402 (0.008\%) measurements classified as extreme outliers. Appendix \ref{app:feature_set} lists the proportion of outliers detected at an aggregated feature level.

At the time of writing, this standardized process of outlier detection and removal is unique to our benchmarking system. In contrast, the public pipeline of \citet{harutyunyan2017multitask} does not perform \emph{any} outlier detection and replacement\footnote{Note in README: ``**Outlier detection is disabled in the current version**'' \url{https://github.com/YerevaNN/mimic3-benchmarks/commit/2da632f0d\#diff-04c6e90faac2675aa89e2176d2eec7d8}}. 
Similarly, the pipeline of \citet{purushotham2018benchmark} does not use outlier removal for its recommended set of 136 raw features, while for their comparison small set of 17 features involved in the SAPS score (including 5 non-time-varying ones) they do remove outliers ``according to medical knowledge'' but provide few reproducible details.
We emphasize that updating the outlier handling of either pipeline would be a labor-intensive change (requiring editing source code).

\subsection*{Hourly Aggregation}
\label{subsec:bucketing}
The raw data in MIMIC-III provides fine-grained timestamps (with resolution in units of seconds or finer) for each laboratory measurement and recorded vital sign.
However, most measurements are infrequent (e.g. blood tests of interest may be run every few hours at most), meaning each variable's raw time-series is quite sparse.
To obtain a denser representation that is easier to reason about and readily applied to modern machine learning methods for time-series that expect discretized time representations, we aggregate the observations from each ICU stay's time-series into hourly buckets.

\subsection*{Semantic Grouping of Raw Features into Clinical Aggregates}
\label{subsec:bucketing}

Each measurements in the MIMIC-III database is associated with a unique ItemID, as specified by the original EHR software. 
These raw ItemIDs are not robust to changes in software or human data entry practices. 
For example, ``HeartRate'' may be recorded under ItemID 211 (using CareVue EHR systems before 2008) or under ItemID 220045 (using MetaVision EHR software after 2008). 
We thus developed a manually curated clinical taxonomy designed to group semantically equivalent ItemIDs together into more robust ``clinical aggregate'' features.
These aggregate representations reduce overall data missingness and the presence of duplicate measures. 
Appendix~\ref{app:feature_set} details the proposed clinical taxonomy about the \texttt{MIMIC-Extract} featurization.
Parallel work by \citet{nestor2019} shows that aggregating via this kind of clinical taxonomy yields significant benefits to the robustness of downstream models with respect to clinical concept drift over time.
Our proposed software pipeline makes this useful taxonomy accessible to researchers and enables reproducibility.

\subsection*{Time-Varying Treatment Labels}
Our code extracts hourly binary indicators of when (if ever) common treatments were provided to each patient over time.
We include \emph{device} treatments such as mechanical ventilation, as well as \emph{drug} treatments such as vasopressors and fluid boluses. 

We target these interventions because they are commonly used in the ICU~\cite{yang1991prospective,mullner2004vasopressors} and, despite medical necessity, they can present notable harms to patients~\cite{tobin2006principles,d2015blood}. We include fluid boluses of two types as interventions, crystalloid and colloid, but do not predict them because they are often considered less aggressive alternatives to vasopressors~\cite{malbrain2014fluid}. The output stores binary indicators of whether an intervention was applied (1) or not applied (0) within a given hour; any missing data is considered a non-treatment (0).

Note that we extract both individual vasopressors (e.g., adenosine, dopamine, norepinephrine, vasopressin, etc.) and overall vasopressor usage, consistent with the MIMIC-III codebase~\cite{johnson2017mimic}. A comprehensive list of extracted interventions is provided in Table~\ref{tab:int_set}.

\begin{table}[h!]
\centering
\resizebox{\linewidth}{!}{%
\begin{tabular}{l|l|r} 
\toprule
\textbf{Intervention}      & \textbf{Concept}                                & \textbf{Mean Hours} \\ 
\midrule
vent               & mechanical ventilation   & 12.20       \\
vaso               & vasopressor              & 8.10        \\
\hspace{3mm}adenosine          & adenosine                & 0.00        \\
\hspace{3mm}dobutamine         & dobutamine               & 0.36        \\
\hspace{3mm}dopamine           & dopamine                 & 0.95        \\
\hspace{3mm}epinephrine        & epinephrine              & 0.60        \\
\hspace{3mm}isuprel            & isuprel                  & 0.01        \\
\hspace{3mm}milrinone          & milrinone                & 0.87        \\
\hspace{3mm}norepinephrine     & norepinephrine           & 2.72        \\
\hspace{3mm}phenylephrine      & phenylephrine            & 4.06        \\
\hspace{3mm}vasopressin        & vasopressin              & 0.90        \\
colloid\_bolus     & colloid bolus            & 0.16        \\
crystalloid\_bolus & crystalloid bolus        & 1.93        \\
nivdurations       & non-invasive ventilation & 25.81       \\
\bottomrule
\end{tabular}}
\caption{Hourly interventions extracted by \texttt{MIMIC-Extract}. Mean Hours is the average number of hours when the continuous interventions are on or when the intermittent interventions (colloid bolus and crystalloid bolus) are administrated, averaged across all patients.
We include separate interventions for 9 distinct vasopressor drugs as well as a general vasopressor intervention when any one is used.
}
\label{tab:int_set}
\end{table}

\subsection*{Extensibility of Data Pipeline}\label{subsec:extensibility}

While \texttt{MIMIC-Extract} promotes reproducibility by providing a default cohort for common benchmark tasks, it is also able to to extract data tailored to specific research questions. In this section, we demonstrate four possible modifications and extensions of this pipeline to enable customized extraction.

\paragraph{Keywords}
Functions in \texttt{MIMIC-Extract} use keywords to control admission cohort and time-varying features selection. Overwriting default values for the following keywords allows researchers to modify default extraction:
\begin{description}
    \item[\texttt{min\_age}] specifies a floor on patients' age to be included in the cohort,
    \item[\texttt{min\_duration} \& \texttt{max\_duration}] specify restrictions on ICU length of stay,
    \item[\texttt{group\_by\_level2}] specifies whether the `raw' or `clinically aggregated' labs and vitals should be extracted, and
    \item[\texttt{min\_percent}] excludes vital and lab variables that contain high proportions of missing values.
\end{description}

\paragraph{Configurable Resource Files}
The extraction code relies on information in associated resource files for variable grouping and extraction (\texttt{itemid\_to\_variable\_map.csv}) and outlier correction (\texttt{variable\_ranges.csv}). By modifying these files, researchers can extract sets of variables that are best suited for specific studies and adjust custom outlier detection thresholds for their task. 
                       
\paragraph{Embedded SQL Queries}
Researchers can modify the code or add SQL queries in the extraction code to include additional static variables, vitals and labs measurements and treatment labels in the output tables. For example, acuity score can be queried and added to the \texttt{patients} table, and treatment fluid amount can be extracted to the \texttt{interventions} table by querying respective tables in the MIMIC relational database. We plan to maintain and update this codebase regularly to reflect additional research needs and improve the extensibility and ease of adding new SQL queries. 

\paragraph{Additional Dataframes}
By using a consistent cohort for all output dataframes, \texttt{MIMIC-Extract} reduces the workload on subsequent data processing in downstream tasks. While it currently extracts static variables, vital signs, lab measurements, and treatment interventions, MIMIC-III contains more clinical information such as prescriptions or diagnostic codes. Researchers can extend the pipeline to output additional groups of variables. This pipeline can also be extended to extracting unstructured data such as caregiver notes to enable multi-modal learning. 

\section*{Comparison to Other Extraction Systems}
\label{sec:comparison}
\begin{table*}[h]
\centering
\resizebox{\linewidth}{!}{%
\begin{tabular}{l|l|c|c|c|c}
\toprule

   &    & \texttt{MIMIC-Extract} & \citet{harutyunyan2017multitask} & \citet{purushotham2018benchmark} & \citet{sjoding2019democratizing} \\
\midrule
\multirow{3}{*}{\shortstack[l]{\textbf{Prediction} \\\textbf{Target}}} 
                            & Mortality              & Y  &  Y &  Y  & Y \\
                            & Length-of-Stay (LOS)   & Y  &  Y &  Y  &   \\
                            & Phenotyping (ICD code) &    &  Y &  Y  &   \\
                            & Physiological Shock    &    &    &     & Y \\
                            & Acute Respiratory Failure (ARF)  &    &    &     & Y \\
                            & Ventilator intervention & Y  &    &     &   \\
                            & Vasopressor intervention & Y  &    &     &   \\
                            & Fluid Bolus intervention & Y  &    &     &   \\
\midrule
\multirow{2}{*}{\shortstack[l]{\textbf{Prediction} \\\textbf{Framework}}} & Fixed Input, Fixed Target    & Y  & Y & Y & Y       \\
                             & Dynamic Input, Dynamic Target  & Y   & Y  &   &      \\
\midrule
\multirow{2}{*}{\textbf{Cohort}}        & Generic        &  Y  &   &    &   \\
                                        & Task-Specific  &     & Y & Y  & Y  \\
\midrule
\multirow{5}{*}{\shortstack[l]{\textbf{Time-Varying} \\\textbf{Feature} \\ \textbf{Representation}}} 
								   & Raw Features                & 269    &  n/a      & 136             & ?  \\
                                   & Clinical Aggregate Features &  104    &   17      &  12 & ?  \\
                                   & Unit Conversion             & Y      &   Y       & not for raw     & ?  \\
                                   & Outlier Detection           & Y      & disabled  & not for raw     & ?     \\
                                   & Missingness Thresholding    & Y      & Y         & Y               & Y \\
\midrule
\multirow{2}{*}{\textbf{Output}} & Format  & .h5 & .csv    & .npy  &  .npz       \\
                                 & Presentation & Cohort  & Patient &  Cohort   &  Cohort       \\
\bottomrule
\end{tabular}}
\caption{Comparisons of public MIMIC-III data pipelines. 
``Y'' indicates a ``yes''.
\citet{purushotham2018benchmark} used clinical aggregation and outlier detection only in their ``Feature Set A'' which only considered the 17 variables (12 time-varying, 5 static) used to calculate SAPS-II risk score. Similarly, these processing steps only apply to a hand-selected set of 17 variables in \citet{harutyunyan2017multitask}.
	Due to limited published resources available about \citet{sjoding2019democratizing}, some features are difficult to assess at present.
}
\label{tab:comp_pipeline}
\end{table*}

\begin{figure}[h!]
  \subfloat[Fixed Input, Fixed Target]{
    \includegraphics[width=0.8\linewidth]{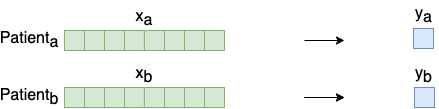}}\\
  \subfloat[Dynamic Input, Dynamic Target]{\includegraphics[width=0.8\linewidth]{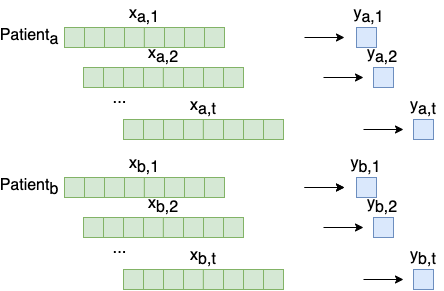}}
  \caption{Common Time-Series Prediction Frameworks. 
  }
  \label{fig:pred_framework}
\end{figure}

A particular reproducibility challenge that the machine learning for health community faces is the lack of standardized data preprocessing and cohort specification \citep{mcdermott2019reproducibility}.
We focus here on the three most similar efforts to ours in addressing this challenge with MIMIC-III: the benchmarks released recently by \citet{harutyunyan2017multitask}, \citet{purushotham2018benchmark}, and \citet{sjoding2019democratizing}. While all these efforts have released public code that transforms MIMIC-III into feature and label sets suitable for supervised machine learning prediction tasks that take multivariate time-series input,
they differ from our work in several important dimensions, including the following:
\begin{itemize}
    \item Prediction Target: Which variables (e.g. mortality, LOS) the task intends to predict.
    \item Prediction Framework: What format input and output data take in the prediction task (see Figure \ref{fig:pred_framework}).
    \item Patient Cohort: Whether the output cohort is generic or task-specific.
    \item Time-varying Feature Representation: What feature representation is chosen for the time-varying variables and what feature transformation is applied.
    \item Output: The format used for output storage and presentation.
\end{itemize}
All works also differ with regard to which patient-specific features are exported and used in prediction, though we do not consider these differences in detail here. Table \ref{tab:comp_pipeline} summarizes the comparison of \texttt{MIMIC-Extract} to these works\footnote{While the comparisons to \citet{harutyunyan2017multitask} and \citet{purushotham2018benchmark} are based on full journal papers, the comparison to \citet{sjoding2019democratizing} is based on a one-page abstract due to publication availability at the time of writing.}. As demonstrated in the comparison, \texttt{MIMIC-Extract} is the only pipeline that generates a generic cohort that can be directly read as Pandas DataFrame. It is also the only pipeline that uses clinical aggregation, unit conversion, and outlier detection on a large set of raw MIMIC-III data.

\paragraph{Prediction Targets}
Mortality and length-of-stay (LOS) are very common targets in relevant benchmark works and are also included in this work. In addition, \texttt{MIMIC-Extract} is the only work demonstrating an intervention prediction task through predicting the onset, offset, stay on, and stay off of mechanical ventilation and vasopressors. This task requires the model to handle the decisions needed in a real ICU where subjects may go on and off treatments throughout their stay using most recently observed data. 

While we do not demonstrate phenotype classification, ICD-9 group classification or acute respiratory failure (ARF) and shock predictions in this work, these prediction targets can be derived either using default \texttt{MIMIC-Extract} output or with slight extensions to the pipeline.


\paragraph{Prediction Framework}
A typical clinical prediction task usually uses one of the two prediction frameworks illustrated in Figure \ref{fig:pred_framework}:
\begin{itemize}
    \item Fixed Input, Fixed Target: A fixed period window of observations is taken from each patient (e.g. the first 24 hours of ICU) and a single target with a fixed temporal relationship to the chosen input window is predicted (e.g. in-hospital mortality or mortality within 30 days).
    \item Dynamic Input, Dynamic Target: Multiple (potentially overlapping)  subsequences are taken from each patient (e.g. the most recent 6 hours). Each input subsequence is used to predict a target variable at a known temporal delay (e.g. remaining LOS, mechanical ventilation onset one hour later). We will consider subsequences of fixed-length in all dynamic benchmarks here (e.g. 6 hour windows), but these could be variable-length in general.
\end{itemize} 

In this work, we profile \texttt{MIMIC-Extract} for both ``Fixed Input, Fixed Target" and ``Dynamic Input, Dynamic Target" frameworks. In particular, we employ three classification tasks: binary mortality prediction (both in-hospital and in-ICU, given the first 24-hour window of patient history), binary long length-of-stay (LOS) prediction (both greater than three and seven days, given the same 24-hour window), and 4-class hourly intervention onset/offset prediction (for both mechanical ventilation and vasopressor administration). 

The only other pipeline work that also handles ``Dynamic Input, Dynamic Target" prediction is by \citet{harutyunyan2017multitask} where they predict hourly next-day mortality, which they call decompensation, and hourly remaining LOS. (Note that they use variable-length subsequences in both tasks, including at each hour information from all previous hours).  However, since they generated task-specific cohorts, it is a more involved process to modify their source code to extract a different dynamic target.

In addition, to the best of our knowledge  \citeauthor{harutyunyan2017multitask}'s work does not employ greater-than-zero gap times when structuring in-hospital mortality prediction, risking temporal leakage of label information when training models. For example, with fixed 48-hour input windows, suppose a patient died at hour 48.5. It is likely that some signals of imminent decline (e.g. last-minute aggressive treatments) would be present before hour 48 and thus included as input, leading the predictor to identify what the care team obviously already knows about the patient's poor health. This is a limitation of these tasks; in this work, all tasks presented use a non-zero temporal gap to ensure no such label leakage.

\paragraph{Cohort}
Our system exports a single cohort, which can be used in a variety of ways under different paradigms for various tasks. In other words, \texttt{MIMIC-Extract}'s raw output does not specify the prediction input features or targets and do not impose task-specific inclusion criteria. All other three works establish either task-specific inclusion criteria or task-specific cohorts with different input features. As a result of producing a generic cohort, \texttt{MIMIC-Extract} is more extensible and easily adapted to different prediction tasks. The absence of task-specific inclusion criteria can also lead to more robust models. Lastly, by focusing more on the general data pipeline workflow rather than task specifics, \texttt{MIMIC-Extract} presents a framework that can be used by machine learning researchers using other clinical datasets.

\paragraph{Time-varying Feature Representation}
Our system exports two possible featurizations: ``raw'' features which match the input representation schema of MIMIC (at the \texttt{ItemID} level), and ``clinical aggregate'' features, where outputs are grouped together according to a manual taxonomy based on clinical knowledge (see Appendix~\ref{app:feature_set}). This representation induces a robustness to underlying temporal concept drift in the representation space \citep{nestor2019}. Note that even though both \citet{harutyunyan2017multitask} and \citet{purushotham2018benchmark} used `processed' feature sets that involves clinical aggregation, unit conversion and outlier detection, they only considered a limited set of features for such transformation. \texttt{MIMIC-Extract} uses these processing techniques to a more comprehensive set of labs and vitals listed in Appendix~\ref{app:feature_set}.

\paragraph{Output}
Other than \citet{harutyunyan2017multitask}, all pipelines generate cohort-level DataFrames or arrays that are easier for data exploration and visualization. \texttt{MIMIC-Extract}'s output can be read directly into a Pandas DataFrame that offers greater readability and easier querying. 

\section*{Output Cohort Characterization} 
\label{sec:cohort}

Our pipeline produces a cohort of 34,472 patients by default with diverse demographic and admission coverage, as summarized in Table~\ref{tab:count}.
Alternative definitions of desired cohort properties (minimum age, etc.) can yield different cohorts. More details about the distribution of various features over this cohort can also be found in Appendix~\ref{app:feature_set}, which details, among other things, the relative rates of missingness for both the individual raw \texttt{ItemID}s and the grouped clinical aggregates over this cohort.


\begin{table*}[h]
\centering
\resizebox{0.6\linewidth}{!}{%
\begin{tabular}{llrrr} 
\toprule
               &                  & \multicolumn{2}{c}{\textbf{Gender}}   & \textbf{Total}   \\
               &                  & F      & M      &         \\ 
\midrule
\textbf{Ethnicity}      & Asian            & 370    & 472    & 842 (2\%)    \\
               & Hispanic  & 448    & 689    & 1,137 (3\%)  \\
               & Black            & 1,448  & 1,219  & 2,667 (8\%)  \\
               & Other            & 2,061  & 3,122  & 5,183 (15\%)  \\
               & White            & 10,651 & 13,992 & 24,643 (71\%) \\ 
\midrule
\textbf{Age}            & \textless{}30    & 748    & 1,084  & 1,832 (5\%)  \\
               & 31-50            & 2,212  & 3,277  & 5,489  (16\%) \\
               & 51-70            & 4,888  & 8,054  & 12,942 (38\%) \\
               & \textgreater{}70 & 7,130  & 7,079  & 14,209 (41\%) \\
\midrule
\textbf{Insurance Type} & Self Pay         & 125    & 352    & 477     (1\%)\\
               & Government       & 402    & 648    & 1,050  (3\%) \\
               & Medicaid         & 1,186  & 1,596  & 2,782  (8\%) \\
               & Private          & 4,415  & 7,431  & 11,846 (34\%) \\
               & Medicare         & 8,850  & 9,467  & 18,317 (53\%) \\ 
\midrule
\textbf{Admission Type} & Urgent           & 409    & 528    & 937    (3\%) \\
               & Elective         & 2,282  & 3,423  & 5,705  (17\%) \\
               & Emergency        & 12,287 & 15,543 & 27,830 (81\%) \\ 
\midrule
\textbf{First Careunit} & TSICU            & 1,777  & 2,725  & 4,502  (13\%) \\
               & CCU              & 2,185  & 3,008  & 5,193  (15\%) \\
               & SICU             & 2,678  & 2,842  & 5,520  (16\%) \\
               & CSRU             & 2,326  & 4,724  & 7,050  (20\%) \\
               & MICU             & 6,012  & 6,195  & 12,207 (35\%) \\ 
\midrule
\textbf{Total}          &                  & 14,978 (43\%) & 19,494 (57\%) & 34,472 (100\%) \\
\bottomrule
\end{tabular}}
\caption{Default Cohort Summary by Static Demographic and Admission Variables.}
\label{tab:count}
\end{table*}

\section*{Benchmark Tasks and Models}
\label{sec:benchmark}
In this section, we profile several benchmark tasks, ranging in complexity, across several types of models using data extracted with \texttt{MIMIC-Extract}, in an effort to both provide meaningful benchmarks and baseline results for the community, as well as to demonstrate the utility of this extraction system. Code to run these benchmarks is available in the form of accompanying Jupyter Notebooks. 

We specifically endeavor to highlight tasks of varying complexity, each with a broad clinical intervention surface. Accordingly, we categorize our benchmarks as two low complexity tasks and one high complexity task. 
Our low complexity tasks are both static, binary classification tasks, each broken into two variants: mortality prediction (either in-hospital or in-ICU) and long length-of-stay (LOS) prediction (either $>3$ day or $>7$ day). Our high complexity task is the hourly prediction of the onset, offset, stay on and stay off of various interventions, as performed in, e.g., \citet{suresh2017clinical}.

Notably, we do not include any tasks based on billing code prediction; while such tasks were included as benchmarks by \citet{harutyunyan2017multitask}, and are commonly used as a target~\cite{lipton2015learning,choi2016retain,choi2018mime}, we argue that predicting diagnosis code is of minimal value clinically, given the lack of temporal association linking a diagnosis to a particular point in the record, and the fact that such codes are more associated with the \emph{billing} of a patient than the \emph{treatment} of said patients~\cite{agniel2018biases}.

We use a non-zero time gap between the most recent feature measurement time and a relevant forecasted event in all tasks. A gap is needed to allow practitioners time to respond to a predicted risk; suddenly warning that a patient is in instant critical need is not viable in medical practice. Additionally, time is needed to assemble care teams or fetch necessary drugs or equipment.

\subsection*{Mortality and Length-of-stay (LOS) Predictions}
\label{sec:simple_baselines}
Risk prediction tasks like mortality and long LOS predictions are highlighted as benchmark tasks in both \citet{purushotham2018benchmark} and \citet{harutyunyan2017multitask}. Though common, they are known to be relatively easy prediction tasks, with performance saturating given only minimal data and even under relatively modest models, such as random forests~\cite{cheRecurrentNeuralNetworks2018,nestor2019}

\subsubsection*{Task Definitions}
We consider several varieties of these tasks, including in-ICU mortality, in-hospital mortality, LOS $> 3$ days prediction, and LOS $> 7$ days prediction. For all tasks, we use clinically grouped time-varying labs and vitals features alone to predict these targets as binary classification task. In all cases, we use the first 24 hours of a patient's data, only considering patients with at least 30 hours of present data. This 6 hour gap time is critical to prevent temporal label leakage, and must be included in any valid benchmark.

\subsubsection*{Data Pre-processing}
Values were mean centered and scaled to unit variance, then missing data was imputed using a variant of the ``Simple Imputation'' scheme outlined in \citet{cheRecurrentNeuralNetworks2018}, in which we represent each variable via a mask (1 if the value is present at this timestep, 0 otherwise), the imputed variable, and the time since the last observation of this feature (with values which have never been observed being given a sentinel large value). In particular, variable values are first forward filled and then set to individual-specific mean if there are no previous values. If the variable is never observed for a patient, its value is set to training set global mean.

\subsubsection*{Models Benchmarked}
For all tasks, we profiled logistic regression (LR), random forest (RF), and gated recurrent unit with delay (GRU-D) \cite{cheRecurrentNeuralNetworks2018} models. As the point of this work is not to make strong statements about the workings or efficacy of these models, but rather to introduce our extraction pipeline and demonstrate its use on benchmark tasks, we will not discuss the details of these models here, but refer the reader to external sources for more model details.


Models were tuned using random hyperparameter search \citep{bergstra2012random} under broad parameter distributions, with 60 hyperparameter samples for RF and LR models, and a variable number of samples for GRU-D (less than 60 in all cases) as GRU-D is significantly more computationally intensive. Note that this likely induces a small bias against GRU-D in these baseline results.


\subsubsection*{Results}
Results for these models are shown in Table~\ref{tab:simple_baselines}. Our AUROCs are very much in line with the literature for these tasks, showing robustly high performance for GRU-D and RF models, as expected. One interesting observation is that random forest models often have poor F1 scores, even while maintaining competitive AUPRC scores. This may indicate that these models are more sensitive to the initial choice of threshold than are other models. Similarly, GRU-D often displays stronger performance under the AUPRC metric than the AUROC metric relative to other models, which likely speaks in its favor here given the strong rates of class imbalance in these tasks.


\begin{table}[h]
    \centering
    \resizebox{\linewidth}{!}{
    \begin{tabular}{llrrrr} \toprule
        Task                                   & Model & AUROC & AUPRC & Accuracy & F1     \\ \midrule
        \multirow{3}{*}{In-ICU Mortality}      & LR    & $88.7$& $46.4$& $93.4\%$ & $38.4$ \\
                                               & RF    & $89.7$& $49.8$& $93.3\%$ & $12.6$ \\
                                               & GRU-D & $89.1$& $50.9$& $94.0\%$ & $43.1$ \\ \midrule
        \multirow{3}{*}{In-Hospital Mortality} & LR    & $85.6$& $49.1$& $91.1\%$ & $42.1$ \\
                                               & RF    & $86.7$& $53.1$& $90.7\%$ & $19.6$ \\
                                               & GRU-D & $87.6$& $53.2$& $91.7\%$ & $44.8$ \\ \midrule
        \multirow{3}{*}{LOS $>$ 3 Days}        & LR    & $71.6$& $65.1$& $68.6\%$ & $59.4$ \\
                                               & RF    & $73.6$& $68.5$& $69.5\%$ & $59.5$ \\
                                               & GRU-D & $73.3$& $68.5$& $68.3\%$ & $62.2$ \\ \midrule
        \multirow{3}{*}{LOS $>$ 7 Days}        & LR    & $72.4$& $18.5$& $91.9\%$ & $7.2$  \\
                                               & RF    & $76.4$& $19.5$& $92.3\%$ & $0.0$  \\
                                               & GRU-D & $71.0$& $17.9$& $91.2\%$ & $10.7$ \\ \bottomrule
    \end{tabular}}
    \caption{Performance Results on In-ICU Mortality, In-Hospital Mortality, $>3$ Day LOS, and $>7$ Day LOS. Classification threshold used for computing accuracy and F1 is set to 0.5.  (Note that due to their additional computational overhead, GRU-D models were undersampled during hyperparameter turning as compared to LR and RF models.)}
    \label{tab:simple_baselines}
\end{table}

\subsection*{Clinical Intervention Prediction}
\label{sec:clin_int_pred}
We also use \texttt{MIMIC-Extract} for intervention prediction tasks. Well-executed intervention prediction can alert caregivers about administrating effective treatments while avoiding unnecessary harms and costs~\cite{wu2016ssam, ghassemi2017predicting}. In a high-paced ICU, such decision-support systems could be a fail-safe against catastrophic errors. We argue that tasks like intervention prediction have a stronger time-series focus and are clinically actionable.
Following prior work on clinical intervention prediction~\cite{suresh2017clinical,ghassemi2017predicting,wu2016ssam}, we present models for predicting two target interventions, mechanical ventilation and vasopressors.

\subsubsection*{Task Definitions}
To make clinically meaningful predictions, we extract from \texttt{MIMIC-Extract} clinically aggregated outputs a sliding window of size 6 hours as input features, then predict intervention onset/offset within a 4 hour prediction window offset from the input window by a 6 hour gap window. For each intervention at each prediction window, there are 4 possible outcomes:

\begin{description}
    \item[Onset] When the intervention begins off and is turned on.
    \item[Stay On] When the intervention begins on and stays on.
    \item[Wean] When the intervention begins on and is stopped.
    \item[Stay Off] When the intervention begins off and stays off.
\end{description}

\subsubsection*{Data Pre-processing}
 Time-varying lab and vital data are preprocessed in a manner similar to that used in the mortality and LOS prediction, except that the ``time since last measure'' column is also centered and rescaled as this is found to improve performance for our neural models. We also include 5 static variables (gender, age bucket, ethnicity, ICU type, and admission type) and time-of-day as additional features. 

\subsubsection*{Models Benchmarked}
We profile LR, RF, convolutional neural network (CNN) models, and Long Short-Term Memory (LSTM) models for this task. Hyperparameters for RF and LR models were tuned via random search, whereas for CNN and LSTM models, parameters were replicated from prior work by \citet{suresh2017clinical}. 

\subsubsection*{Results}
Model performance is summarized in Table~\ref{tab:int_results}.

\begin{table}[h]
\centering
\resizebox{\linewidth}{!}{%
\begin{tabular}{l|rr|rr|rr|rr} \toprule
          & \multicolumn{2}{c|}{RF} & \multicolumn{2}{c|}{LR} & \multicolumn{2}{c|}{CNN} & \multicolumn{2}{c}{LSTM} \\ 
          & Vent. & Vaso.          & Vent. & Vaso.          & Vent. & Vaso.           & Vent. & Vaso.            \\ \midrule
Onset AUROC     & 87.1  & 71.6 & 71.9 &   68.4& 72.2  & 69.4  & 70.1  & 71.9             \\
Wean AUROC     & 94.0  & 94.2 & 93.2 &   93.9 & 93.9  & 94.0  & 93.1  & 93.9            \\
Stay On AUROC  & 98.5  & 98.5 & 98.4 &  98.2   & 98.6  & 98.4 & 98.3  & 98.3             \\
Stay Off AUROC & 99.0  & 98.3  & 98.3  &   98.5  & 98.4  & 98.1  & 98.4  & 98.1             \\ \midrule
Macro AUROC & 94.6  & 90.7 & 90.4  &  89.8   & 90.8  & 90.0  & 90.0  & 90.1             \\ \midrule
Accuracy & 79.7  & 83.8 & 78.5  &  72.9   & 61.8  & 77.6  & 84.3  &     82.6         \\ \midrule
Macro F1 & 48.1  & 48.9 & 47.7  &  45.1   & 44.4  & 44.4  & 50.1  &         48.1     \\ \midrule
Macro AUPRC & 42.7  & 42.0 & 43.1  &  40.2   & 42.4  & 38.9  & 44.4  &          41.7    \\ 
\bottomrule
\end{tabular}}
\caption{Performance Results on Mechanical Ventilation and Vasopressor Prediction.}
\label{tab:int_results}
\end{table}

We find that CNN and LSTM models perform very similarly to prior studies---this is notable given we \emph{do not} include notes, whereas many prior studies do \citep{suresh2017clinical}. RF models perform surprisingly well, outperforming CNN and LSTM models and prior results reported in the literature.

\section*{Design Choices and Limitations}

While \texttt{MIMIC-Extract} aims to be flexible in supporting a wide range of machine learning projects using MIMIC-III, we make several design choices that may render \texttt{MIMIC-Extract} less relevant to tasks that differ significantly from the benchmark tasks presented in this paper. 

Most notable among these designed choices are the features we exclude. Notable such categories include prescriptions, certain labs and vitals, various treatments/interventions, and notes. Many of these features can be externally extracted and joined to our pipeline's output (as we demonstrate in `Extensibility of Data Pipeline' Section for notes), and others we exclude intentionally due to concerns about their robustness (prescriptions), but other parties may wish to extend the pipeline to enable extraction of these features. 

In addition, our time-series coarsening into hourly buckets can also be limiting for certain tasks. By bucketing data into hourly aggregates, we lose out on a level of granularity present in the raw data and force the irregular medical timeseries into a artificially regular representation. We also lose all granularity with regards to time-of-day, which has known effects on care delivery~\cite{agniel2018biases}. Similarly, our clinical groupings, while highly performant, are also manually curated and limit the extensibility of the pipeline to new labs and vitals.

\section*{Conclusion} \label{sec:conclusion}

\texttt{MIMIC-Extract} is an open source cohort selection and pre-processing pipeline for obtaining multivariate time-series for clinical prediction tasks. 
The system produces a single, large cohort and represents time-varying data according to manually-defined, clinically meaningful groupings. This representation shows strong performance and robustness to care practice drift.
We demonstrate that this pipeline can be used in a diverse range of clinical prediction tasks. We hope its focus on usability, reproducibility, and extensibility will help spur development of machine learning methodology via clinically relevant and reproducible benchmark tasks.
Ultimately, we hope \texttt{MIMIC-Extract} will enable easier and faster development of effective machine learning models that might drive improvements in delivering critical care.

\section*{Code availability}

The full \texttt{MIMIC-EXTRACT} pipeline code, including SQL queries and configurable resource files, as well as Jupyter Notebooks walking through benchmark tasks and models are available at \url{https://github.com/MLforHealth/MIMIC_Extract}.

\section*{Acknowledgments}

Matthew McDermott is funded in part by National Institutes of Health: National Institutes of Mental Health grant P50-MH106933 as well as a Mitacs Globalink Research Award. Geeticka Chauhan acknowledges the support of the Wistron Corporation in Taiwan. Dr. Marzyeh Ghassemi is funded in part by Microsoft Research, a CIFAR AI Chair at the Vector Institute, a Canada Research Council Chair, and an NSERC Discovery Grant. Dr. Michael C. Hughes acknowledges support in part from NSF Projects HDR-1934553 and IIS-1908617.

\bibliographystyle{ACM-Reference-Format}
\bibliography{main}

\newpage
\appendix
\onecolumn
\clearpage

\begin{landscape}
\section{Feature Set}
\label{app:feature_set}
\begin{small}
Columns ``low'', ``high'', and ``strict'' indicate the proportion of observations corrected under outlier detection. ``NAN'' indicates that there is no prescribed valid range for that variable.
Columns ``avg.'', ``std.'', and ``pres.'' indicate the arithmetic mean, the standard deviation, and the percentage of present (non-missing) values for each variable.
Columns ``pres. cv'' and ``pres. mv'' are the proportion of present (non-missing) values from patients whose data are recorded under Carevue and Metavision software systems, respectively.
\end{small}
\begin{tiny}
\begin{longtable}{l r r r r r r r r r r r r r r}

    \hline \hline \multicolumn{1}{c}{\textbf{Grouping}} & \multicolumn{1}{c}{\textbf{low}}
    & \multicolumn{1}{c}{\textbf{high}}& \multicolumn{1}{c}{\textbf{strict}}& \multicolumn{1}{c}{\textbf{avg}} & \multicolumn{1}{c}{\textbf{std}} & \multicolumn{1}{c}{\textbf{pres.}} &
    \multicolumn{1}{c}{\textbf{pres. cv}} &
    \multicolumn{1}{c}{\textbf{pres. mv}} &\multicolumn{1}{c}{\textbf{ItemID}} & \multicolumn{1}{c}{\textbf{Table}} &
    \multicolumn{1}{c}{\textbf{DB}} &\multicolumn{1}{c}{\textbf{avg}} &
    \multicolumn{1}{c}{\textbf{std}} &\multicolumn{1}{c}{\textbf{pres.}}\\ \hline 
    \endfirsthead
    
    \hline \multicolumn{1}{c}{\textbf{Grouping}} & \multicolumn{1}{c}{\textbf{low}}
    & \multicolumn{1}{c}{\textbf{high}}& \multicolumn{1}{c}{\textbf{strict}}& \multicolumn{1}{c}{\textbf{avg}} & \multicolumn{1}{c}{\textbf{std}} & \multicolumn{1}{c}{\textbf{pres.}} &
    \multicolumn{1}{c}{\textbf{pres. cv}} &
    \multicolumn{1}{c}{\textbf{pres. mv}} &\multicolumn{1}{c}{\textbf{ItemID}} & \multicolumn{1}{c}{\textbf{Table}} &
    \multicolumn{1}{c}{\textbf{DB}} &\multicolumn{1}{c}{\textbf{avg}} &
    \multicolumn{1}{c}{\textbf{std}} &\multicolumn{1}{c}{\textbf{pres.}}\\ \hline 
    \endhead
    
    \endfoot
    
    \hline \hline
    \endlastfoot

    \hline
            \multirow{3}{3.8cm}{alanine aminotransferase}    & \multirow{3}{0.9cm}{1.7E-04}    & \multirow{3}{0.9cm}{1.9E-04} & \multirow{3}{0.9cm}{3.9E-04}    & \multirow{3}{0.9cm}{282.3}    & \multirow{3}{0.9cm}{916.4} & \multirow{3}{0.9cm}{2.0E-02} & \multirow{3}{0.9cm}{2.7E-02} & \multirow{3}{0.9cm}{1.5E-02}& 50861 & labevents & lab & 278.6 & 909.6 & 1.5E-02\\

										&&&&&&&&& 769    & chartevents & cv  & 335.1 & 984.9 & 6.7E-03\\

										&&&&&&&&& 220644    & chartevents & mv  & 281.1 & 907.0 & 6.7E-03\\

    \hline
            \multirow{4}{3.8cm}{albumin}    & \multirow{4}{0.9cm}{0.0E+00}    & \multirow{4}{0.9cm}{1.9E-05} & \multirow{4}{0.9cm}{0.0E+00}    & \multirow{4}{0.9cm}{3.1}    & \multirow{4}{0.9cm}{0.7} & \multirow{4}{0.9cm}{1.3E-02} & \multirow{4}{0.9cm}{1.7E-02} & \multirow{4}{0.9cm}{1.1E-02}& 50862 & labevents & lab & 3.1 & 0.7 & 1.0E-02\\

										&&&&&&&&& 772    & chartevents & cv  & 2.9 & 0.6 & 4.9E-03\\

										&&&&&&&&& 1521    & chartevents & cv  & 3.0 & 0.6 & 3.8E-03\\

										&&&&&&&&& 227456    & chartevents & mv  & 3.1 & 0.7 & 4.1E-03\\

    \hline
            \multirow{1}{3.8cm}{albumin ascites}    & \multirow{1}{0.9cm}{NAN}    & \multirow{1}{0.9cm}{NAN} & \multirow{1}{0.9cm}{NAN}    & \multirow{1}{0.9cm}{1.7}    & \multirow{1}{0.9cm}{0.8} & \multirow{1}{0.9cm}{5.2E-05} & \multirow{1}{0.9cm}{5.8E-05} & \multirow{1}{0.9cm}{4.5E-05}& 50835 & labevents & lab & 1.7 & 0.8 & 5.2E-05\\
\end{longtable}
\end{tiny}
\end{landscape}

\end{document}